\documentclass[10pt,twocolumn,letterpaper]{article}

\usepackage{wacv}      

\usepackage{graphicx}
\usepackage{amsmath}
\usepackage{amssymb}
\usepackage{booktabs}

\usepackage[pagebackref,breaklinks,colorlinks]{hyperref}

\usepackage[capitalize]{cleveref}
\crefname{section}{Sec.}{Secs.}
\Crefname{section}{Section}{Sections}
\Crefname{table}{Table}{Tables}
\crefname{table}{Tab.}{Tabs.}

\def\wacvPaperID{2006} 
\def\confName{WACV}
\def\confYear{2024}

\newcommand{\cliset}{\mathcal{C}}
\newcommand{\trndata}{D_T}
\newcommand{\valdata}{D_S}

\newcommand{\flmdl}{f_{\theta}}

\newcommand{\selclients}{\mathcal{C}_{sel}}
\newcommand{\cliupdate}{\delta^t_i}

\newcommand*{\argmin}{\mathop{\mathrm{argmin}}}

\newcommand{\model}{\mbox{\textsc{Gcfl}}} 
\newcommand{\EE}{\mathbb{E}}
\newcommand{\noniid}{\emph{non-i.i.d.\ }}
\newcommand{\pgt}{\Pr_{S}}

\newcommand{\coreset}{\Xcal_i^t}

\newcommand{\ompsubset}{\mathcal{G}}

\newcommand{\param}{\theta}

\newcommand{\cliloss}{\ell_i}
\newcommand{\serverloss}{\ell_S}

\newcommand{\corewt}{{\mathbf w}}
\newcommand{\clilossgm}{\ell_i^{gm}}

\usepackage{amssymb}
\usepackage{mathtools}
\usepackage{subcaption}
\usepackage{multirow}
\usepackage{enumitem}

\usepackage{todonotes}
\usepackage{xcolor} 
\usepackage{tikz}
\usetikzlibrary{fit,calc}
\newcommand*{\tikzmk}[1]{\tikz[remember picture,overlay,] \node (#1) {};\ignorespaces}
\newcommand{\boxit}[1]{\tikz[remember picture,overlay]{\node[yshift=3pt,fill=#1,opacity=.25,fit={(A)($(B)+(.95\linewidth,.8\baselineskip)$)}] {};}\ignorespaces}
\definecolor{cyan}{rgb}{0.0, 0.75, 1.0}
\definecolor{pink}{rgb}{1.0, 0.8, 0.6}

\usepackage{algorithm}
\usepackage{algcompatible}
\usepackage{adjustbox}

\algnewcommand\algorithmicreturn{\textbf{return}}
\algnewcommand\RETURN{\State \algorithmicreturn}%
\usepackage{def}

\begin{document}

\title{Gradient Coreset for Federated Learning}

\author{
Durga Sivasubramanian$^\dagger$\\
{IIT Bombay}\\
{\tt\small durgas@cse.iitb.ac.in}
\and
Lokesh Nagalapatti$^\dagger$\\
{IIT Bombay}\\
{\tt\small nlokeshiisc@gmail.com}
\and
Rishabh Iyer\\
{University of Texas at Dallas}\\
{\tt\small Rishabh.Iyer@utdallas.edu}
\and
Ganesh Ramakrishnan\\
{IIT Bombay}\\
{\tt\small ganesh@cse.iitb.ac.in}
}
\maketitle
\def\thefootnote{$\dagger$}\footnotetext{L.N. and D.S. contributed equally}\def\thefootnote{\arabic{footnote}}
\begin{abstract}
   Federated Learning (FL) is used to learn machine learning models with data that is partitioned across multiple clients, including resource-constrained edge devices. It is therefore important to devise solutions that are efficient in terms of compute, communication, and energy consumption, while ensuring compliance with the FL framework's privacy requirements. Conventional approaches to these problems select a weighted subset of the training dataset, known as coreset, and learn by fitting models on it. Such coreset selection approaches are also known to be robust to data noise. However, these approaches rely on the overall statistics of the training data and are not easily extendable to the FL setup.

    In this paper, we propose an algorithm called \mbox{\textsc G}radient based \mbox{\textsc C}oreset for Robust and Efficient \mbox{\textsc F}ederated \mbox{\textsc L}earning (\model) that selects a coreset at each client, only every $K$ communication rounds and derives updates only from it, assuming the availability of a small validation dataset at the server. We demonstrate that our coreset selection technique is highly effective in accounting for noise in clients' data. We conduct experiments using four real-world datasets and show that \model{} is (1) more compute and energy efficient than FL, (2) robust to various kinds of noise in both the feature space and labels, (3) preserves the privacy of the validation dataset, and (4) introduces a small communication overhead but achieves significant gains in performance, particularly in cases when the clients' data is noisy. 
\end{abstract}

\vspace{-1cm}
\section{Introduction}
\label{introduction}

Federated learning (FL) is an approach to machine learning in which clients collaborate to optimize a common objective without centralizing data~\cite{mcmahan2016communication}. The training dataset is distributed across a group of clients, and they contribute to the training process by sharing privacy-preserving updates with the central server across communication rounds until the model converges.

FL proves particularly valuable in situations where a central server lacks a sufficient amount of data for standalone model training but can leverage the collective data from multiple clients, including edge devices, sensors, or hospitals. For instance, a hospital aiming to develop a cancer prediction model can benefit from training their model using data from other hospitals, while maintaining the privacy of sensitive patient information. However, in scenarios where clients have limited computational resources or their data is noisy, it becomes imperative to design robust algorithms that enable their participation while minimizing computation and energy requirements. Our proposed solution addresses this challenge by identifying a subset of each client's data, referred to as the "coreset," which reduces noise and facilitates effective training of the central server's model.

\begin{figure}
    \centering
    \includegraphics[width=0.5\textwidth]{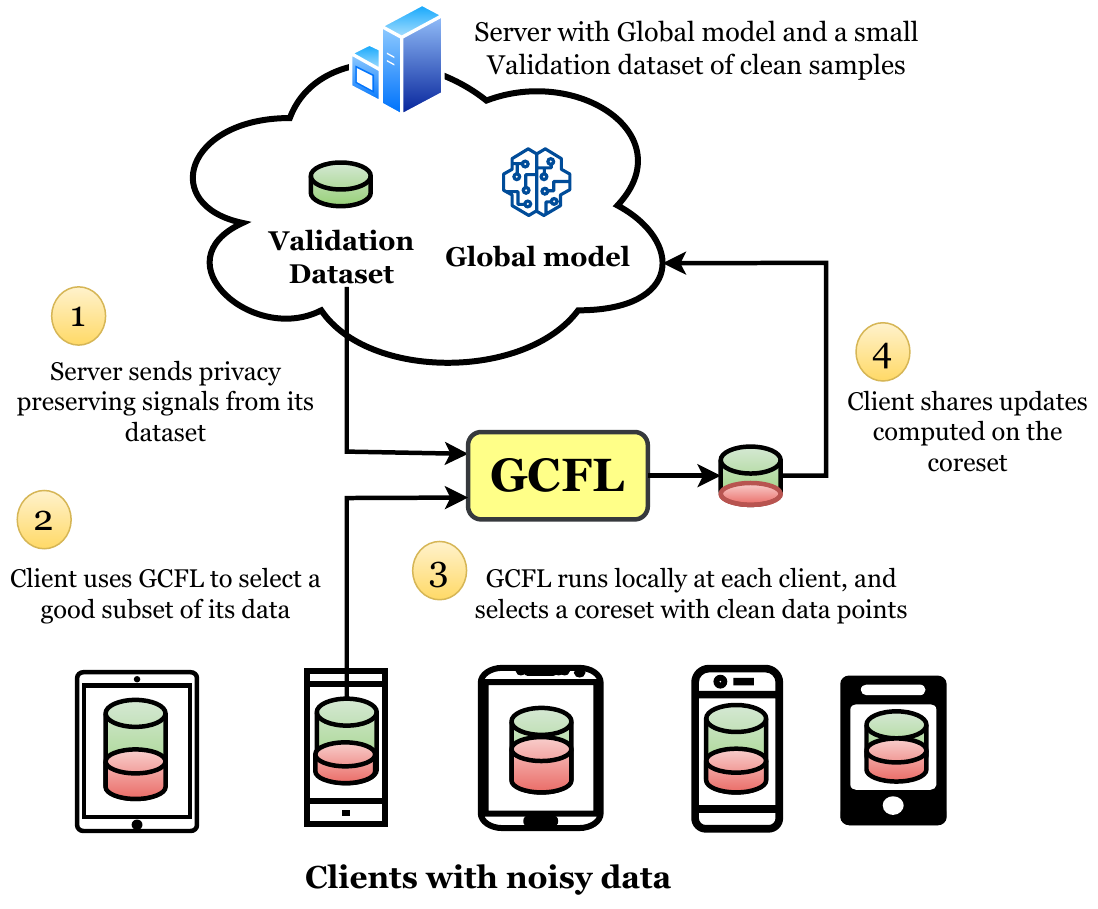}
    \caption{Schematic overview of \model. We illustrates a server with a limited validation dataset and multiple participating clients, which are edge devices with data that contain noise. 
    }
    \vspace{-0.5cm}
    \label{fig:teaser}
\end{figure}

Conventional coreset selection methods such as Facility Location, CRUST, CRAIG, Glister, and Gradmatch~\cite{crust, facilitylocation, craig, glister} have been developed to enhance the efficiency and robustness of machine learning model training. However, adapting these strategies to Federated Learning (FL) settings is challenging due to the non-i.i.d. nature of clients' datasets. Traditional coreset algorithms aim to select a representative subset that ensures a model trained on it performs similarly to the entire dataset. In FL, each client's data originates from diverse distributions and is influenced by noise. For example, in a hospital context, data distributions differ due to varying demographics across locations and may be subject to varying amounts of different types of noise. As a result, biased updates from clients hinder the learning progress of FL algorithms. We demonstrate this obstacle through a motivating experiment in Section \ref{sec:motivation}.

We introduce \model{} (as shown in Figure ~\ref{fig:teaser}), an algorithm specifically designed to address the aforementioned challenges. Our approach involves selecting a coreset every $K$ communication rounds to derive local updates. Similar to \cite{fednoniid}, we assume the server has access to a small validation dataset for guiding coreset selection. However, in contrast to \cite{fednoniid}, our validation dataset is not public; we exclusively use (last layer) gradients derived from it. \model{} uses these gradients to identify a coreset at each client, effectively training the FL model. In our experiments, we demonstrate that our approach is robust to different types of noise, efficient, while preserving privacy and minimizing communication overhead\footnote{The code can be found at  \url{https://github.com/nlokeshiisc/GCFL_Release/tree/master}}.

In summary, our work makes the following contributions:

\begin{enumerate}[wide, labelwidth=!, labelindent=0pt]
\itemsep 0em
\item We introduce \model, a framework for efficiently selecting coresets for federated learning while preserving privacy.
\item Through experiments, we show that \model\ achieves the best tradeoff between accuracy and speed in a non-noisy setting.
\item Furthermore, we demonstrate that \model\ effectively filters out various types of noise, including closed-set label noise, open-set label noise, and attribute noise, resulting in improved performance compared to well-established baselines for FL and coreset selection.
\end{enumerate}

\section{Motivating experiment}\label{sec:motivation}

To emphasize the need for a coreset algorithm like \model{} in federated learning and to showcase its impact on performance, we conducted an experiment using a small toy dataset. We generated ten isotropic Gaussian blobs in $\mathbb{R}^{10}$ with varying standard deviations (ranging from $1$ to $8$) using scikit-learn's $\texttt{make\_blob()}$ utility~\cite{scikit-learn}. A test set was reserved, containing $15\%$ of the samples, while the remaining training data was divided among the server and ten clients. The training subset allocated to the server serves as a validation dataset, as will be explained in our algorithm, and is solely employed to guide coreset selection at the clients.

\begin{figure}[t]
    \centering
    \includegraphics[width=0.8\linewidth]{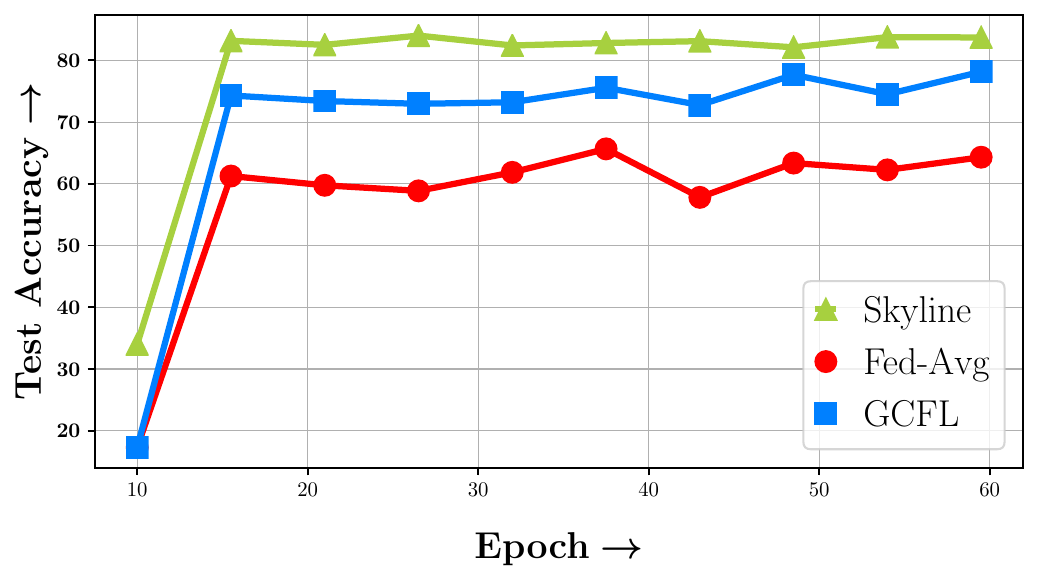}
    \caption{Performance of FedAvg, \model, and skyline under 40\% label noise. Skyline is trained just on the clean points. \model\ performs comparably to the skyline.} 
    \label{fig:motiv}
\end{figure}

To simulate noise, we randomly flipped $40\%$ of the labels in each client's samples. We trained logistic regression models under three settings: (i) FedAvg, (ii) \model{}, and (iii) Skyline. In the Skyline approach, clients only computed updates from the clean ($60\%$) samples to establish an upper-bound performance benchmark using clean data. Figure~\ref{fig:motiv} displays the results, revealing that FedAvg's performance is adversely affected by the presence of noisy training samples. In contrast, \model{} outperforms FedAvg and falls between Skyline, demonstrating its effectiveness in mitigating the impact of noisy FL data, likely due to its use of a small, server-guided training subset. \model{} holds significant promise for FL, especially in noisy data settings, where it significantly enhances accuracy and model generalization.

One can consider the idea of mitigating the impact of noise by fine-tuning the model refined from each communication round using the server's validation dataset. However, such an approach may prove ineffective when the sample size in the validation dataset is insufficient. To explore this approach, we conducted experiments using CIFAR-10 and CIFAR-100 datasets, both affected by $40\%$ closed-set label noise. We compared the performance of \model{} and FedAvg, as detailed in Table \ref{tab:mov_fine_tune}. Our observations indicate that while fine-tuning does offer some improvements, its effectiveness is limited by the small size of the validation dataset. In our upcoming experiments, we will demonstrate how \model{}, in contrast, efficiently harnesses the validation dataset to guide coreset selection at the client side, ultimately optimizing its performance.

\begin{table}
  \begin{center}
    {\small{
\begin{tabular}{ccccc}
\toprule
Dataset & FedAvg & FedAvg + & \model{} & \model{} +\\
& & Fine Tuned &  & Fine Tuned \\

\midrule
Cifar10  & 34.1\% & 34.6 \% & 47.4 \% & 49.5 \% \\
Cifar100 & 11.6\% & 12.1 \% & 17.5\% & 17.9 \% \\
\bottomrule

\end{tabular}
}}
\end{center}
\caption{Impact of model fine-tuning at the server with $D_S$ under 40\% noise. Fine-tuning yields minor enhancements, while using $D_S$ for coreset selection results in substantial improvements.}
\label{tab:mov_fine_tune}
\end{table}

\section{Related Work}

Federated Learning (FL) is distinguished by data heterogeneity, where various clients possess data from different sources with diverse characteristics. As a result, aggregating updates from such a heterogeneous data source can impact the convergence rate of models. The challenge of addressing this heterogeneity within client data in the context of Federated Learning (FL) has garnered significant attention in the literature \cite{mmfed1, mmfed2, fedem, fedprox, scaffold, moon}.

For instance, FedProx, introduced by \cite{fedprox}, incorporates a proximal term into the objective function. This term penalizes updates that deviate significantly from the server's parameters, aiming to accommodate client heterogeneity. Similarly, Scaffold, proposed by \cite{scaffold}, focuses on reducing the variance in the server's aggregated updates by managing the drift in update computation across clients. However, FL presents a unique challenge when dealing with noisy data owned by clients, as neither the server nor the clients have a complete view of the entire training dataset. Traditional data cleaning approaches \cite{fednoniid2, fednoniid3, fednoniid4, fedrobustnoniid} may not be directly applicable to FL.

Coreset selection is a well-established technique in machine learning that involves selecting a weighted subset of data to approximate a desired quantity across the entire dataset. Traditional coreset selection methods typically depend on the model and use submodular proxy functions for coreset selection \cite{submodcoreset1, submodcoreset2, submodcoreset3, coresetkmeans, submodcoreset4}. Recent developments in the literature have explored coreset selection in conjunction with deep learning models \cite{craig, crust, gradmatch, deepcoreset1, facilitylocation}. Nevertheless, most existing coreset selection approaches are tailored for conventional settings where all data is readily accessible, requiring thoughtful adaptation for FL.

Coreset selection in Federated Learning (FL) is an underexplored domain, primarily due to the complexities tied to privacy and non-i.i.d. data distribution among clients \cite{clientsel1, clientsel2, clientsel4, dvrl_fed}. Notably, \cite{diverseblimes} picks a coreset of clients to collectively represent the global update across all clients using the facility location algorithm. In contrast, \cite{clientsel1} uses Shapley values in a game-theoretic framework for again to perform client selection, while \cite{dvrl_fed} explores reinforcement learning techniques for data selection. However, Nonetheless, training \cite{dvrl_fed} is a challenging task, imposing an additional workload on local clients by necessitating the training of an extra private model.  In comparison to the prior work, \model{} is easy to implement and blends well with the FL framework.

FL has different paradigms: Personalized FL strives to train specialized models for individual clients, and substantial research has been conducted in this direction \cite{collins2021exploiting, FL_MAML, pmlr-v162-marfoq22a, Li_2022_CVPR, NEURIPS2021_f8580959}. In contrast, our work is focused on building models that exclusively account for the server's distribution.

We finally note that various techniques have been introduced to ensure privacy in FL, including differential privacy \cite{diff_privacy_1, diff_privacy_2}, homomorphic encryption \cite{homomorphic_1, homomorphic_2}, and more. As \model\ is model-agnostic, these methods can be seamlessly integrated with our approach.

\section{Problem Setup}
In our Federated Learning setup, a group of $N$ clients is represented by the set $\cliset = \{c_1,c_2,\cdots, c_N\}$. The training dataset $\trndata = \bigcup_{i=1}^N D_i$ is divided among the clients, where each client $c_i$ has a data chunk $D_i$ consisting of $n_i$ samples $\{(x_{ij}, y_{ij})\}_{j=1}^{n_i}$. Here, $x_{ij} \in \mathcal{X}$ denotes the input features of the $j^{th}$ data point at the $i^{th}$ client, and $y_{ij} \in \mathcal{Y}$ represents its corresponding target. It is important to note that the data chunks are disjoint, and the set of samples in each data chunk is \textbf{not} obtained independently and identically distributed ($\iid$) from the ground truth target distribution $\pgt$.

Our objective is to train a machine learning model $\flmdl: \mathcal{X} \rightarrow{\mathcal{Y}}$ where $\theta$ represents the learnable parameters. The server $S$ defines the objective for the downstream task and has access to a small dataset $\valdata$ consisting of samples obtained independently and identically from the ground truth target distribution $\pgt$. Our aim is to minimize the expected value of the loss function $\ell(\flmdl(x), y)$, over instances $(x, y)$ sampled from the distribution $\pgt$. 

\vspace{-0.4cm}
\begin{align}
\min_{\theta} \EE_{(x, y) \sim \pgt(\bullet, \bullet)}\big[\ell(\flmdl(x), y)\big] \label{eq:mainobj}
\end{align}

As $\valdata$ is small, it is insufficient for training $\flmdl$ and using it alone can lead to overfitting. To overcome this, the server seeks assistance from the clients to learn $\flmdl$ while respecting their privacy constraints. Federated Learning is a promising solution to this problem, where the learning progresses through $T$ communication rounds. In each round $t$, the server selects a subset of clients $\selclients^t$ and shares the current FL model parameters $\theta^t$ with them. The selected clients initialize their local model with $\theta^t$ and train it for a few epochs with their respective private data chunks $D_i$ to arrive at the updated model parameters $\theta_i^\prime$. The difference in model parameters, computed as $\delta_i^t = \theta^\prime_i - \theta^t$, is then transmitted back to the server. The server then averages the parameter updates received from clients and updates the FL model as follows:

\begin{align}
    \theta^{t+1} &= \theta^t + \eta_g \frac{1}{|\selclients^t|} \sum\limits_{i\in\selclients^t} \cliupdate \label{eq:fedavg}
\end{align}

The global learning rate used by the server is denoted as $\eta_g$, while $\eta_l$ represents the local learning rate used by each client to train $\model{}$. Although updates generated using equation~\eqref{eq:fedavg} can minimize the objective~\eqref{eq:mainobj} when the clients' datasets are independently and identically distributed according to $\pgt$, computing updates from the entire dataset is not recommended when the data contains noise. Moreover, for resource-constrained clients such as edge devices, it is crucial to compute updates in an energy-efficient manner. To address these challenges, we propose using adaptive coreset selection, which involves selecting a weighted subset that approximates the characteristics of the entire dataset. The selected coreset should reduce computation costs without compromising the performance of the FL model, and also prevent the updates from only minimizing the client's local loss, especially when the client's data distribution significantly differs from the ground truth distribution $\pgt$. In this regard, we aim to answer the following question: How can clients in $\cliset$ select an effective coreset that facilitates the computation of $\cliupdate$ and also helps minimize the objective~\eqref{eq:mainobj}?

\section{The \model{} Solution Approach}

Let us denote the coreset selected by a client $c_i$ in communication round $t$ as $\coreset$ and its associated weight as $\corewt_i^t$. We begin the exposition by listing certain desiderata for coreset selection in FL and then proceed to explain how \model meets them.

\begin{enumerate}[wide, labelwidth=!, labelindent=0pt]
\itemsep 0em
    \item The algorithm should align with the current data distribution of clients while also approximating the ground truth distribution $\pgt$, guaranteeing a coreset that mirrors the desired target.
    \item The coreset algorithm should adapt and update $\coreset$ as FL model $\flmdl$ evolves, maintaining relevance.
    \item Assumptions in the coreset approach should uphold FL privacy constraints, safeguarding client data and confidentiality.
\end{enumerate}

To meet the first requirement, relying solely on signals from a client's local dataset, denoted as $D_i$, is insufficient, as these datasets are not identically and independently distributed with respect to the global distribution $\pgt$. However, $\valdata$ contains samples drawn from the target distribution and can potentially aid in the selection of a coreset. Previous research~\cite{fednoniid} has demonstrated that making $\valdata$ publicly accessible enables clients to choose an effective coreset. Nevertheless, in privacy-sensitive domains like healthcare, even the inclusion of $\valdata$ may raise privacy concerns, making it unsuitable for sharing.

Hence, the task of coreset selection within the input feature space becomes challenging. As an alternative, we shift our focus towards coreset selection in the gradient space, as Federated Learning (FL) allows the server to disseminate gradients computed from $\valdata$. It is important to note that any cryptographic techniques applied to secure clients' gradients, such as differential privacy, can also be employed to safeguard the server's gradients. Additionally, to minimize communication overhead, we opt to transmit an aggregated gradient (average) derived from $\valdata$. This choice is grounded in the Information Bottleneck theory~\cite{info_bottle_2}, which suggests that parameters in the final layers contain crucial discriminatory class information $\mathcal{Y}$, while those in the initial layers primarily encapsulate feature-specific information $\mathcal{X}$.

We begin \model\ by defining the server's objective and then illustrate how we incorporate it within the Federated Learning framework. We use $\serverloss$ to signify the loss incurred by the server concerning the validation data $\valdata$, and denote the loss of client $c_i$ w.r.t. its local dataset $D_i$ as $\cliloss$.

\begin{align}
\serverloss &= \frac{1}{|\valdata|}\sum_{(x, y) \in \valdata} \ell(\flmdl(x), y) \label{eq:serverloss} \\
\cliloss &= \frac{1}{n_i} \sum_{(x, y) \in D_i} \ell(\flmdl(x), y) \label{eq:cliloss}
\end{align}
where $\ell: \mathcal{Y} \times \mathcal{Y} \rightarrow \mathbb{R}^+$ is a loss function that is pertinent to the problem.

We define $\nabla_\theta \serverloss(\theta)$ as the average gradient of $\serverloss$ at $\theta$ on the validation dataset $\valdata$, and 
$\{\nabla_\theta \cliloss^j\}_{j=1}^{n_i}$ 
as individual data gradients of client $c_i$ for all $j \in [n_i], i \in [N]$. The objective for each client $c_i$ is to select a coreset $\coreset, \corewt_i^t$ of size $b$ such that the gradients derived from it closely match $\nabla\theta\serverloss(\theta)$. Our coreset selection objective is:

\begin{align}
    \underset{{\Xcal_i^t \subseteq D_i \text{ s.t. } |\Xcal_i^t| \leq b}}{\operatorname{argmin\hspace{0.7mm}}} 
    \min_{\corewt_i^t} \mbox{E}_\lambda(\mathbf{w}_i^t, \Xcal_i^t) \text{ where, } \label{eq:gmobj}
\end{align}
\begin{align}
     \mbox{E}_\lambda(\mathbf{w}_i^t, &\Xcal_i^t) = \lambda \lVert \mathbf{w}_i^t \rVert^2 + {\big\Vert \sum_{j \in \Xcal_i^t} w^t_{ij} \nabla_{\theta}\ell_i^j(\theta^t) -  \nabla_{\theta}\serverloss (\theta^t)\big\Vert}  \nonumber
\end{align}
Here, $\lambda$ is a hyper-parameter that regulates the weights of selected items in the coreset. Due to the combinatorial nature of the optimization objective, it is known to be NP-Hard \cite{glister}. Therefore, we employ a greedy approximation method, which we will explain in more detail later.

Assuming that the client has solved Eq. \ref{eq:gmobj}, we now describe how it computes an update to share with the server. Let $\clilossgm$ denote the loss on the selected coreset, defined as
\begin{align}
\clilossgm = \sum_{j \in \coreset}  \cliloss^j(\param) \label{eq:clilossgm}
\end{align}
To minimize the above loss, the client runs several epochs of stochastic gradient descent on the coreset. From the updated model, the client derives its update $\cliupdate$ as follows:
\begin{align}
\cliupdate = \theta^{t} - \frac{\eta_l}{b} \sum_{k=1}^E \sum_{j \in \Xcal_i^t} \nabla_{\theta} \cliloss^j(\theta^t_{k-1}) \label{eq:cliupdate}
\end{align}
Where $E$ is the number of local gradient update steps performed by the client on the coreset, and $\theta^t_{k-1}$ denotes the model parameters at the $k^{th}$ intermediate step, with $\theta^t_{0} = \theta^t$. In our experiments, we observed that the coreset weight $\corewt_i$ had a minimal impact on computing the update, so we omitted it in Eq ~\eqref{eq:cliupdate}.

\begin{figure}
    \centering
    \includegraphics[width=8cm,height=6.5cm]{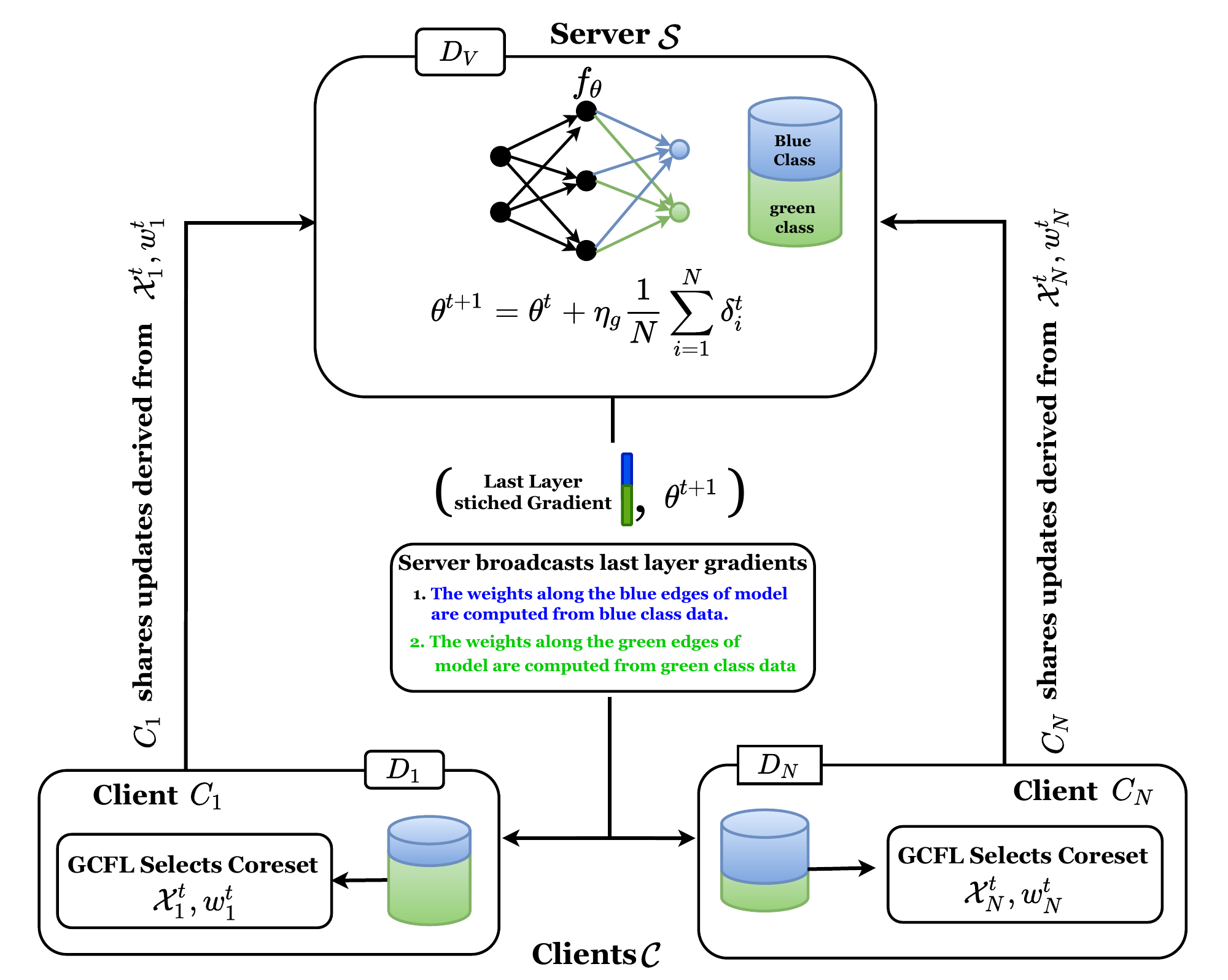}
    \caption{This demonstrates the workflow of \model{} for binary classification with {\color{blue}blue} and {\color{green}green} classes. The server transmits the final layer gradients from the validation dataset $\valdata$. The client employs the OMP algorithm to select a coreset $\Xcal_i^t, w_i^t$, which is used to compute updates shared with the server.}
    \label{fig:arch}
\end{figure}

\subsection{Greedy solution to select Coreset~\eqref{eq:gmobj}}
Objective~\eqref{eq:gmobj} presents a challenging combinatorial optimization problem due to the discrete variable $\coreset$. However, if $\coreset$ is fixed, the inner optimization problem over weights $\corewt_i^t$ can be addressed using the Orthogonal Matching Pursuit (OMP) algorithm, as also used in ~\cite{gradmatch}. Here's a detailed algorithm description:

The coreset selection algorithm operates iteratively, selecting points sequentially until the budget is exhausted. To illustrate, let's consider adding the $(k+1)^{\text{th}}$ point while assuming that $k$ points have already been selected. We denote the coreset with $k$ points as $\ompsubset_i^k$ and its associated weights as $\corewt_i^k$.

At this stage, the choice of the data point, denoted as $j \in \{[n_i] - \ompsubset_i^k\}$, for inclusion in $\ompsubset_i^{k+1}$ is made based on minimizing the error residue. The residue, denoted as $r^k$, is computed as follows:
$r^k = \sum_{j \in \ompsubset_i^k} \corewt_{ij}^k \nabla_\theta \cliloss^j(\theta^t) - r^{k-1}$
Here, $r^0$ is initialized using the server's broadcasted validation gradient $\theta^t$. The purpose of $r^k$ is to quantify the remaining error that needs reduction through the addition of more points to the coreset.

We choose $j$ as $\underset{j \in [n_i] \text{ s.t. } j\not \in \ompsubset_i^{k}}{\argmin} {\big\Vert \nabla_{\theta}\ell_i^j(\theta^t) -r^{k}\big\Vert}$, i.e., the data point that minimizes the distance between its gradient and the residue. The residue's norm monotonically decreases with the coreset's size increase. The pseudocode for the greedy selection is avaiable in Alg. \ref{alg:greedy} and for the overall algorithm in Alg. \ref{alg:grad_match} in the Appendix.

Next, we discuss techniques to help clients implement the greedy algorithm effectively. 
In practice, clients can employ heuristics to avoid solving $b$ iterations to select a coreset of size $b$ by selecting multiple data points at each greedy iteration. Such an approach, backed by strong approximation guarantees, reduces the frequency of running the greedy algorithm. 

To reduce computational overhead, we use the Information Bottleneck theory from~\cite{info_bottle_2}. This theory shows that the initial layers of deep neural networks capture input distribution, while later layers hold task-specific data. In \model, the server only transmits the gradient from the softmax layer, which guides the greedy algorithm in selecting data subsets that minimize the softmax layer's error. This strategic selection significantly trims computational costs, maintains accuracy, and reduces communication expenses by transmitting only a fraction of gradients. Moreover, the computational burden of the greedy algorithm is lessened due to the reduced dimensionality of the OMP problem.

\subsection{Label-wise Coreset Selection} Here, we introduce an improved version of \model{} for better alignment with Federated Learning. Given the data's \noniid\ nature, clients often hold imbalanced class label distributions among their samples \cite{fedclassimb1, fedclassimb2}. Hence, a per-class coreset selection by clients is desirable. The server broadcasts $|\mathcal{Y}|$ gradients, each corresponding to a distinct class $y' \in \mathcal{Y}$ and derived from loss on samples $\{(x, y) \in \valdata| y = y'\}$. Subsequently, we execute $|\mathcal{Y}|$ instances of the greedy algorithm, each selecting a coreset of approximately size $\frac{b}{|\mathcal{Y}|}$. Importantly, this strategy does not increase the computational overhead as the number of greedy iterations remains fixed. Furthermore, the number of gradients per Linear Regression instance within the greedy algorithm diminishes to about $\frac{b}{|\mathcal{Y}|}$, which leads to a reduction in the computational requirements. However, broadcasting the server's gradient increases communication costs by a factor of $|\mathcal{Y}|$. In the following section, we delve into a simple fix to alleviate this issue.

\subsection{Broadcasting Label-wise gradients}
To minimize communication costs, we leverage the idea that when conducting coreset selection for a particular class $y\in \mathcal{Y}$, the server only needs to transmit gradients related to the penultimate layer's connection with the output neuron for class $y$. This approach retains the original gradient broadcast size. The label-wise coreset variant significantly trims computational expenses while maintaining communication efficiency. We present a pictorial overview of the label-wise coreset selection variant in Figure \ref{fig:arch}.

\section{Experiments}

\begin{figure*}[t]
\centering
\includegraphics[width =\textwidth, trim={0cm .5cm 0 0},clip] {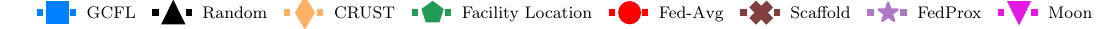}
\centering
\begin{subfigure}[b]{0.22\textwidth}
\centering
\includegraphics[width=\textwidth]{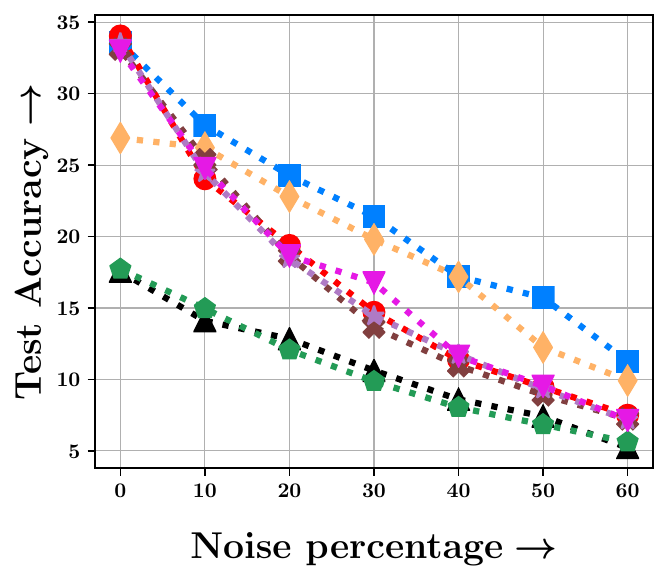}
\caption*{$\underbracket[1pt][1.0mm]{\hspace{3.8cm}}_{\substack{\vspace{-4.0mm}\\
\colorbox{white}{(a) \scriptsize CIFAR100}}}$}
\phantomcaption
\label{fig:CIFAR100}
\end{subfigure}
\begin{subfigure}[b]{0.22\textwidth}
\centering
\includegraphics[width=\textwidth]{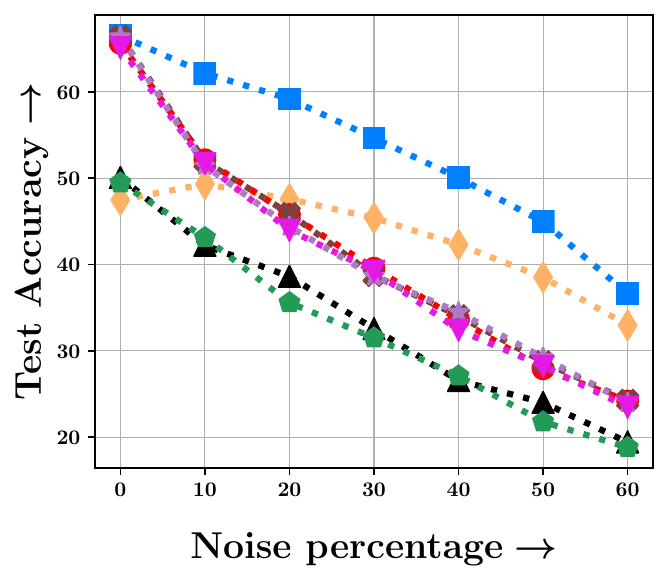}
\caption*{$\underbracket[1pt][1.0mm]{\hspace{3.8cm}}_{\substack{\vspace{-4.0mm}\\
\colorbox{white}{(b) \scriptsize CIFAR10}}}$}
\phantomcaption
\label{fig:CIFAR10}
\end{subfigure}
\begin{subfigure}[b]{0.22\textwidth}
\centering
\includegraphics[width=\textwidth]{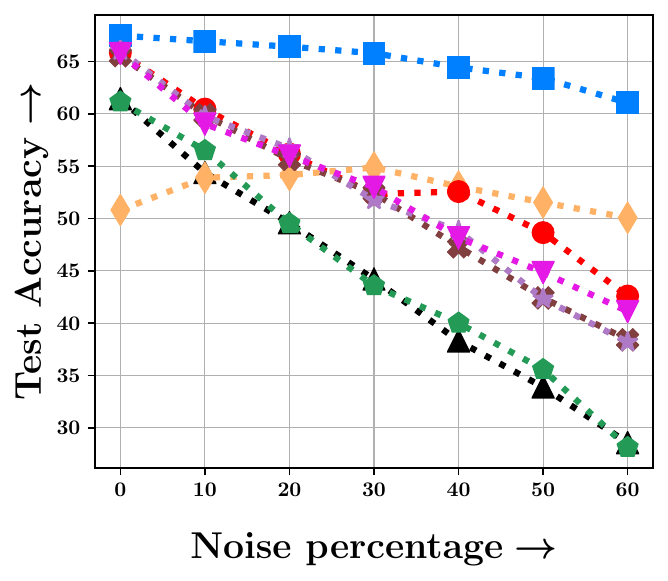}
\caption*{$\underbracket[1pt][1.0mm]{\hspace{3.8cm}}_{\substack{\vspace{-4.0mm}\\
 \colorbox{white}{(c) \scriptsize FEMNIST}}}$}
 \phantomcaption
\label{fig:FEMNIST}
\end{subfigure}
\begin{subfigure}[b]{0.22\textwidth}
\centering
\includegraphics[width=\textwidth]{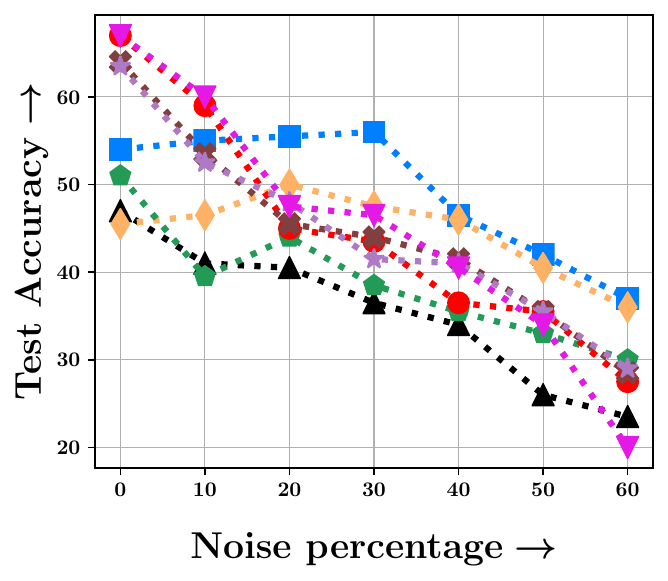}
\caption*{$\underbracket[1pt][1.0mm]{\hspace{3.8cm}}_{\substack{\vspace{-4.0mm}\\
\colorbox{white}{(d) \scriptsize FLOWERS}}}$}
\phantomcaption
\label{fig:Flowers}
\end{subfigure}

\caption{Performance comparison of \model\ and baselines with varying closed-set noise percentages. The X-axis indicates the introduced noise level, and the Y-axis shows test set accuracy. Notably, at x=0, no noise is present. Overall, \model\ outperforms the baselines, except for the flowers dataset, where subset selection hurts.}
\label{fig:coreset_closed}
\end{figure*}

\begin{figure}[t]
\centering
\begin{subfigure}[b]{0.45\linewidth}
\centering
\includegraphics[width=\textwidth]{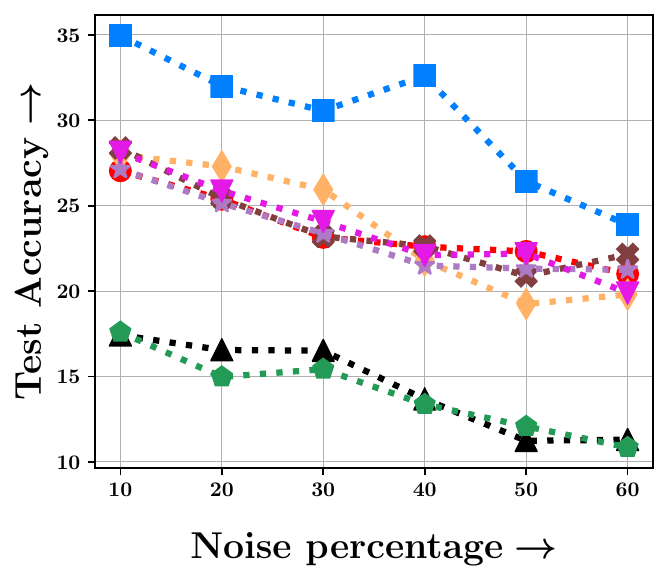}
\caption*{$\underbracket[1pt][1.0mm]{\hspace{3.5cm}}_{\substack{\vspace{-4.0mm}\\
\colorbox{white}{(a) \scriptsize CIFAR100}}}$}
\phantomcaption
\end{subfigure}
\begin{subfigure}[b]{0.45\linewidth}
\centering
\includegraphics[width=\textwidth] {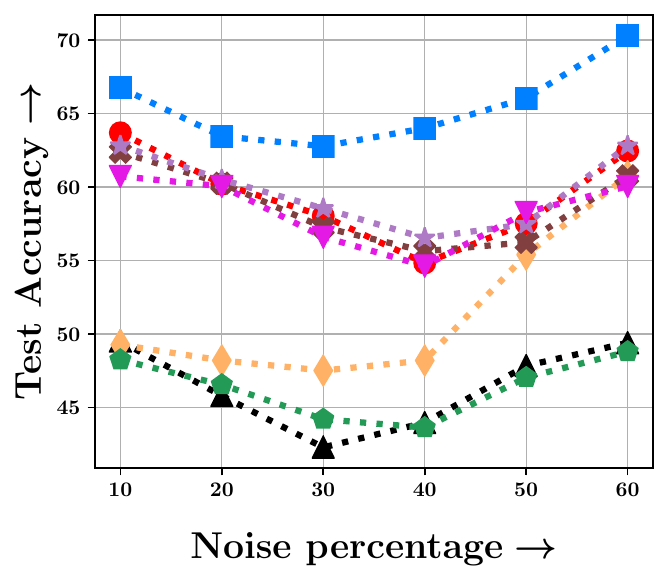}
\caption*{$\underbracket[1pt][1.0mm]{\hspace{3.5cm}}_{\substack{\vspace{-4.0mm}\\
\colorbox{white}{(b) \scriptsize CIFAR10}}}$}
\phantomcaption
\end{subfigure}
\caption{Performance of \model\ in presence of open set noise with 10\% data subset. The legend is borrowed from the Fig \ref{fig:coreset_closed}.}
\label{fig:coreset_open}
\end{figure}

\begin{figure}[t]
\centering
\begin{subfigure}[b]{0.45\linewidth}
\centering
\includegraphics[width=\textwidth]{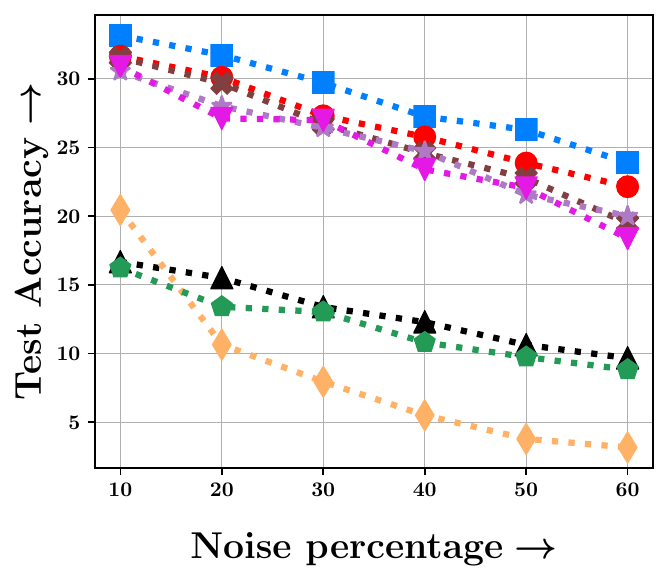}
\caption*{$\underbracket[1pt][1.0mm]{\hspace{3.5cm}}_{\substack{\vspace{-4.0mm}\\
 \colorbox{white}{(a) \scriptsize CIFAR100}}}$}
 \phantomcaption
\end{subfigure}
\begin{subfigure}[b]{0.45\linewidth}
\centering
\includegraphics[width=\textwidth]{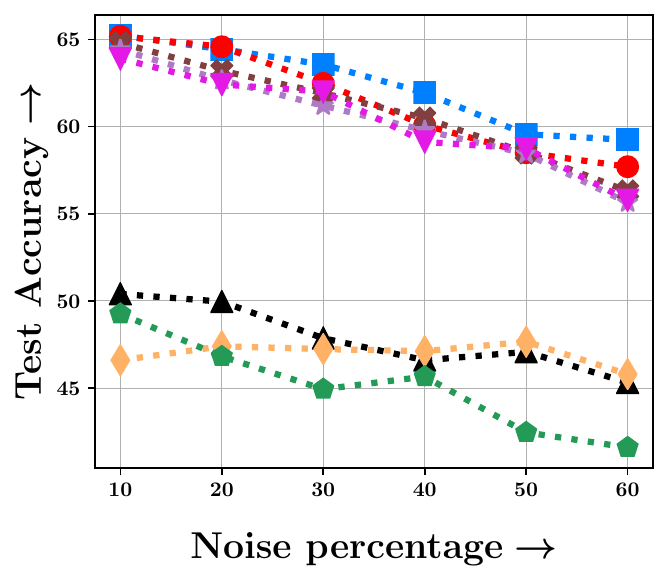}
\caption*{$\underbracket[1pt][1.0mm]{\hspace{3.5cm}}_{\substack{\vspace{-4.0mm}\\
\colorbox{white}{(b) \scriptsize CIFAR10}}}$}
\phantomcaption
\end{subfigure}
\caption{Performance of \model\ in presence of attribute noise with budget $b=10\%$. The legend is borrowed from the Fig \ref{fig:coreset_closed}.\label{fig:coreset_attribute}}

\end{figure}

We present experiments on various real world datasets with a range of noise settings to demonstrate the efficacy of \model{} over state of the art approaches. 

\subsection{Datasets} \label{dataset}
We use four datasets: CIFAR-10, CIFAR-100 ~\cite{Krizhevsky09learningmultiple}, Flowers \footnote{flowers: \url{https://www.tensorflow.org/datasets/catalog/tf_flowers}}, and FEMNIST\cite{caldas2018leaf}, detailed in the appendix. To replicate real-world federated learning scenarios and introduce dataset heterogeneity, we follow prior \noniid\ setups ~\cite{feddirichlet, fedem}. Using a Dirichlet distribution ($\alpha=0.4$), we sample class proportions for clients, distributing data accordingly. We introduce attribute and label noise to showcase \model's robustness. \noniid\ split's impact on dataset heterogeneity is illustrated in Figure \ref{fig:clinetDis}(in the appendix), revealing uneven class and noise distributions across clients.

\subsection{Baselines}
We experiment with two kinds of baselines: coreset baselines that strive to train models with a subset of the training dataset and standard FL algorithms. \\
\noindent\textbf{Coreset selection baselines}:
\begin{enumerate}[wide, labelwidth=!, labelindent=0pt]
\itemsep 0em
    \item \textit{Random} baseline that selects the subset randomly. 
    \item \textit{Facility location} ~\cite{kaushal2019learning}  that  selects a representative subset by maximising the similarity with the ground set.
    \item  \textit{CRUST}~\cite{crust}   a recent coreset based approach to perform robust learning in noisy settings. This could be thought of as an application of \cite{diverseblimes} in our setting. 
\end{enumerate}
\textbf{FL Algorithms with Full Dataset Updates:}
\begin{enumerate}[wide, labelwidth=!, labelindent=0pt]
\addtocounter{enumi}{3}
\itemsep 0em
    \item \textit{Fed-Avg} ~\cite{mcmahan2016communication} The popular FL algorithm that simply averages the updates and applies to the model. 
    \item  \textit{FedProx}~\cite{li2020federated} Controls client drift by introducing $L_2$ regularization to encourage proximity to server parameters.
    \item  \textit{Scaffold}~\cite{scaffold} Reduces variance among client updates to control drift.
    \item  \textit{MOON} ~\cite{moon} Uses contrastive learning to align client and server representations.
\end{enumerate}

\subsection{Model Architecture and Experimental Setup}
We use the SGD optimizer with initial learning rates of $\eta_l = \eta_g = 0.01$, a momentum of $0.9$, and weight decay of $5e-4$. We employ cosine annealing~\cite{loshchilov2017sgdr} to change the learning rate. The server's model architecture consists of a two-layer CNN followed by two fully connected layers. We train models for $T=250$ communication rounds. During each round, coreset-based approaches process only the selected subset. We use a batch size of $32$.

Our experiments demonstrate results that evaluate \model's robustness and efficiency. The robustness analysis compares \model\ with baselines under different noise settings, while the efficiency evaluation considers aspects like computational and communication overhead. Ablation studies further explore \model's sensitivity to different hyperparameters.

\subsection{Robustness}\label{sec:robust} 

Standard Federated Learning methods generally struggle with noisy client data, as evident in our synthetic experiment (Figure \ref{fig:motiv}). In this section, we empirically compare the robustness of \model{} with various FL/coreset algorithms by experimenting with different noise types that exist both in attributes ($\mathcal{X}$) and labels ($\mathcal{Y}$) and assesses their impact.

\noindent\textbf{Closed Set Label noise ~\cite{label_noise_type}: } Closed-set label noise occurs when labels in the training data are incorrect, yet they belong to the true label set $\mathcal{Y}$. To simulate this noise with a ratio of $n\%$, we randomly choose $n\%$ of samples from each $D_i$ and flip their labels. Results for closed-set noise are shown in Figure \ref{fig:coreset_closed}. Notably, \model{} performs the best, particularly with higher noise ratios across different datasets. On the Flowers dataset, \model{} is slightly behind Fed-Avg at lower percentages due to the dataset's small size (only $3670$ images), this dip aligns with other coreset algorithms as well.
        
\noindent\textbf{Open Set Label Noise: ~\cite{label_noise_type}: }
Open-set label noise involves incorrect labels not belonging to the task's label set. To simulate this with a noise ratio $n$, we randomly mark $n\%$ of labels from $\mathcal{Y}$ as irrelevant. This transforms the classification task to focus on the remaining $(1-n)\%$ labels. We retain noisy-labeled features, but adjust their labels by flipping them to other $(1-n)\%$ classes, altering the task to this reduced set.  Figure \ref{fig:coreset_open} illustrates the impact of open-set label noise on \model{} and coreset baselines. We observe that, except for \model, other coreset baselines perform worse than FedAvg baselines, primarily because identifying noisy samples is challenging without guidance from the server. For CIFAR-10, the performance improves as the percentage of open-set noise increases, when $n > 40\%$. This is due to the reduced class count, simplifying the classification task.

\noindent\textbf{Attribute noise ~\cite{attrnoise}: } In contrast to label noise, attribute noise involves corruption of instance features. We use nine types of noise from a library\footnote{https://github.com/bethgelab/imagecorruptions} to corrupt features. Figure \ref{fig:coreset_attribute} shows the effects of attribute noise. We find the Federated Learning models are relatively resilient to attribute noise. This is perhaps due to the data augmentation behavior exhibited with this kind of noise. However coreset selection methods (except \model{}) struggle because detecting noise in attribute space without server guidance is challenging, paralleling the observation with open-set label noise.
   
\begin{figure}[t]
\centering
\includegraphics[width = .75\linewidth,trim={0 0.55cm 0 0},clip] {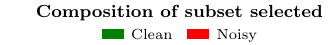}
\centering
\hspace{-0.6cm}
\begin{subfigure}[b]{0.45\linewidth}
\centering
\includegraphics[width=\linewidth]{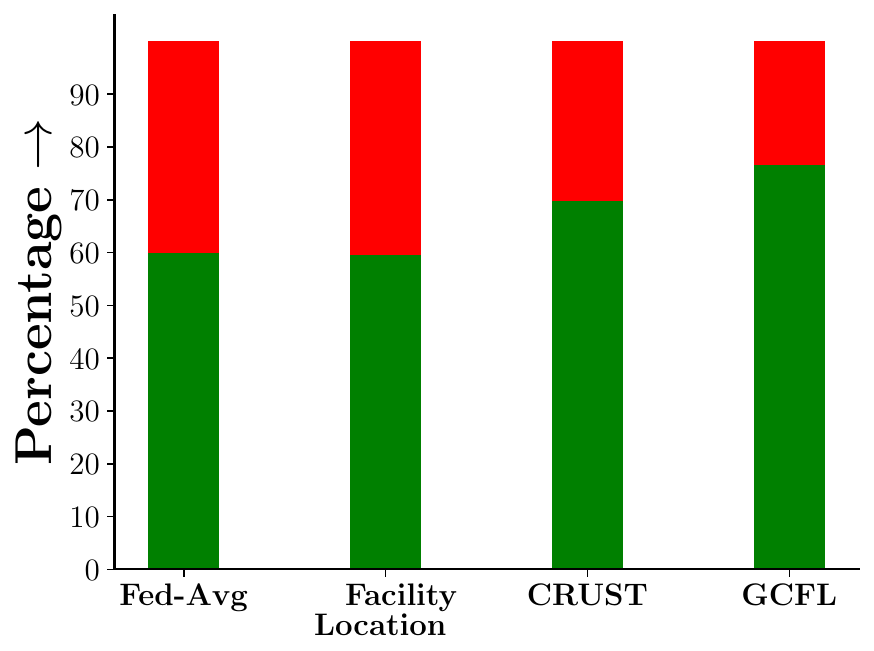}
\caption*{$\underbracket[1pt][1.0mm]{\hspace{3.5cm}}_{\substack{\vspace{-4.0mm}\\
\colorbox{white}{(a) \scriptsize CIFAR-10}}}$}
\phantomcaption
\label{fig:CIFAR100}

\end{subfigure}
\begin{subfigure}[b]{0.45\linewidth}
\centering
\includegraphics[width=\linewidth]{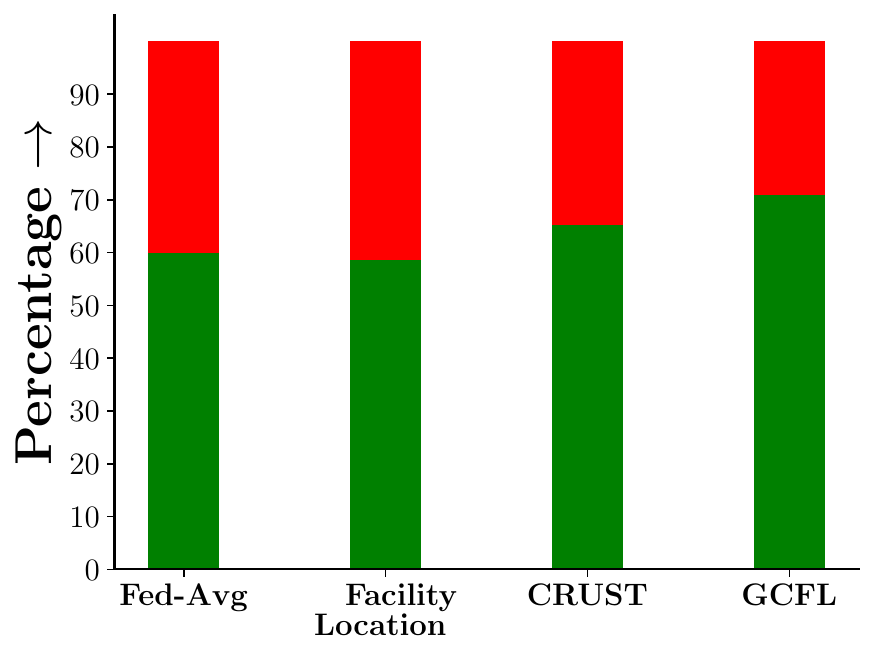}
\caption*{$\underbracket[1pt][1mm]{\hspace{3.5cm}}_{\substack{\vspace{-4.0mm}\\
\colorbox{white}{(b) \scriptsize CIFAR-100}}}$}
\phantomcaption
\label{fig:CIFAR10}
\end{subfigure}
\caption{Here, we examine the number of clean points chosen for the coreset by different subset selection algorithms when trained with 40\% closed-set noise. Notably, \model\ stands out by including a substantial amount of clean points in the coreset.}
\label{fig:composition}
\end{figure} 



It is evident that \model{} outperforms other algorithms in all noise settings, as it can effectively leverage the global information providemaind by the server's guidance to select an appropriate coreset. To further understand the reasons behind \model's superior performance, we conduct a small probing study that is outlined next.
 
\subsection{Does \model{} select clean data points?}
In this experiment, we analyze the composition of subsets selected by various coreset selection algorithms to assess the quality of data points chosen by \model{}. Figure \ref{fig:composition} illustrates the noise composition in the coresets selected for CIFAR-10 and CIFAR-100 datasets under $40\%$ closed-set label noise. The findings demonstrate that among the subset selection algorithms, \model{} chooses the subset with the highest count of clean points. This observation aligns with the improved performance demonstrated in Figure \ref{fig:coreset_closed}. Next, we will discuss the efficiency aspect of \model{} and then proceed to a series of ablation studies.

\begin{figure}[t]
\centering
\includegraphics[width = \linewidth] {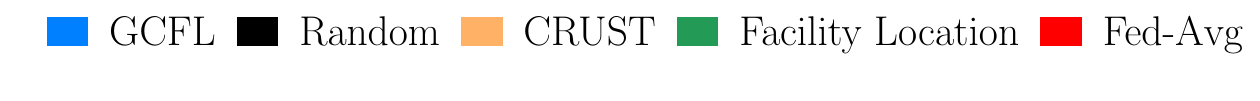}
\centering
\hspace{-0.6cm}
\begin{subfigure}[b]{0.49\linewidth}
\centering
\includegraphics[width=\linewidth]{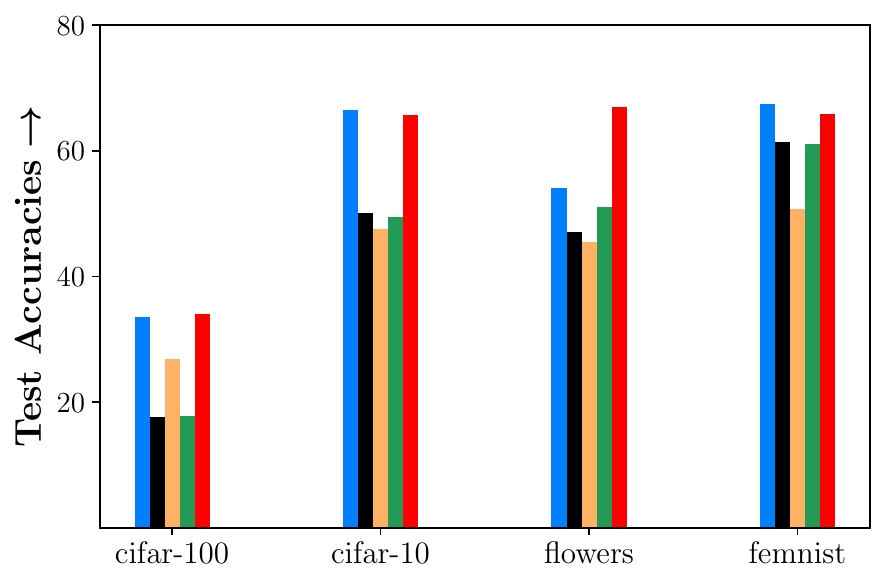}
\caption*{$\underbracket[1pt][1.0mm]{\hspace{4.0cm}}_{\substack{\vspace{-4.0mm}\\
\colorbox{white}{(a) \scriptsize Test Accuracies}}}$}
\phantomcaption
\label{fig:CIFAR100}

\end{subfigure}
\begin{subfigure}[b]{0.49\linewidth}
\centering
\includegraphics[width=\linewidth]{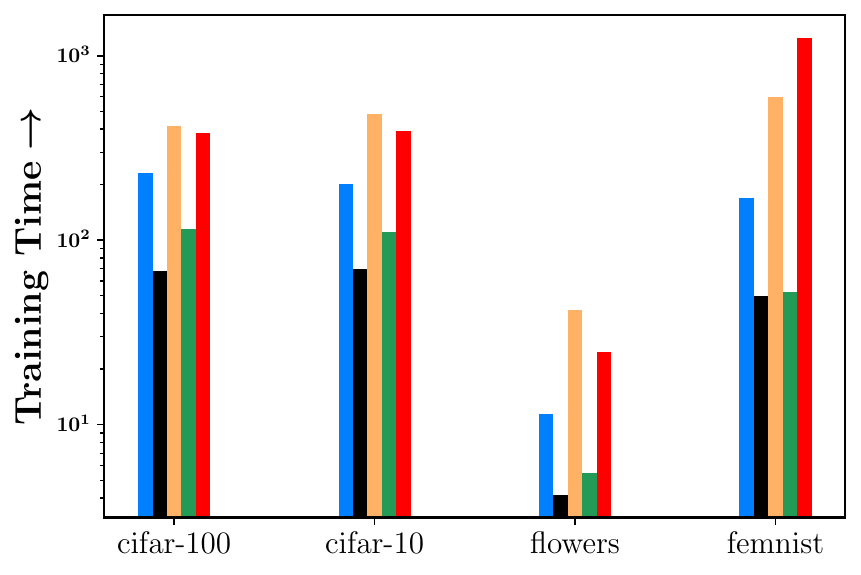}
\caption*{$\underbracket[1pt][1mm]{\hspace{4.0cm}}_{\substack{\vspace{-4.0mm}\\
\colorbox{white}{(b) \scriptsize Timing Analysis}}}$}
\phantomcaption
\label{fig:CIFAR10}
\end{subfigure}
\caption{Trade-off between the training time and test accuracy on the raw datasets without any noise. We set a budget of $b=10\%$.}
\label{fig:eff_experiments}
\end{figure} 

\subsection{Efficiency}\label{sec:eff} 

Reducing the computational cost often involves training the models only on a subset of the dataset. The more informative the subset is, better the performance of the model. We evaluate this in Figure \ref{fig:eff_experiments}, where we compare models trained on $10\%$ coresets selected using different algorithms: Random, Facility Location, CRUST, and \model{}. Due to the \noniid nature of clients datasets in FL, algorithms like CRUST and Facility Location struggle to optimize the global server objective. Moreover, adapting them to FL setup is challenging due to data privacy. \model{}, however, aligns with the server's last-layer loss gradient, resulting in superior performance, except for the small Flowers dataset. Figure \ref{fig:eff_experiments} demonstrates \model{} achieves a compelling accuracy-efficiency trade-off by just selecting $10\%$ coresets every $10$ rounds.

\subsection{Computational overhead of \model{}} 
We conducted a timing analysis on the CIFAR-10 dataset to assess \model's computational overhead. FedAvg consistently takes 1.5 seconds per round. However, \model\ requires 5.6 seconds every $K^{\text{th}}$ round, where coreset selection is performed. In every other round, it incurs only 0.2 seconds. Therefore, for any $K\geq 5$, we would achieve significant computational benefits compared to FedAvg. For instance, with $K=10$, an \model\ epoch averages only 0.76 seconds, using just 50\% of the compute compared to FedAvg.

\subsection{Communication overhead of \model}
\model\ introduces minimal communication overhead in practice, even though it requires transmission of validation set gradients from the server. In our experiments, the server already broadcasts the entire model with around 3.5 million parameters, while the last layer comprises only about 200 thousand parameters. As \model\ operates every $K$ epochs (e.g., $K=10$ in our experiment), the long-term effect results in an additional communication overhead of merely 20 thousand parameters. This amounts to a modest $0.25\%$ increase in communication cost, keeping \model's communication cost practically equivalent to FedAvg.

\subsection{Is GCFL privacy compliant?}
\model\ introduces only one additional component compared to the FedAvg, where the server broadcasts updates on $\valdata$ to assist clients with corset selection. However, our approach requires just the softmax layer gradients. Although previous research has shown that training features can be inferred from individual data gradients, reconstructing samples with just the softmax layer gradients, particularly when averaged across instances, is exceedingly difficult. Therefore, \model{} satisfies the privacy constraints of federated learning.

\begin{figure}[t]
\centering
\includegraphics[width =0.20\linewidth] {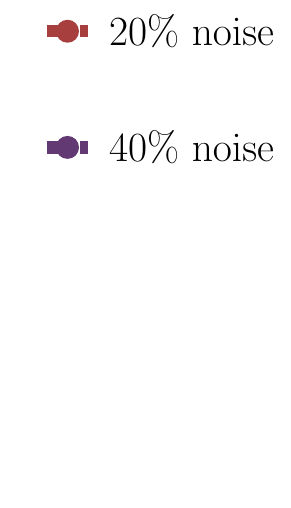}
\centering
\begin{subfigure}[b]{0.39\linewidth}
\centering
\includegraphics[width=\linewidth]{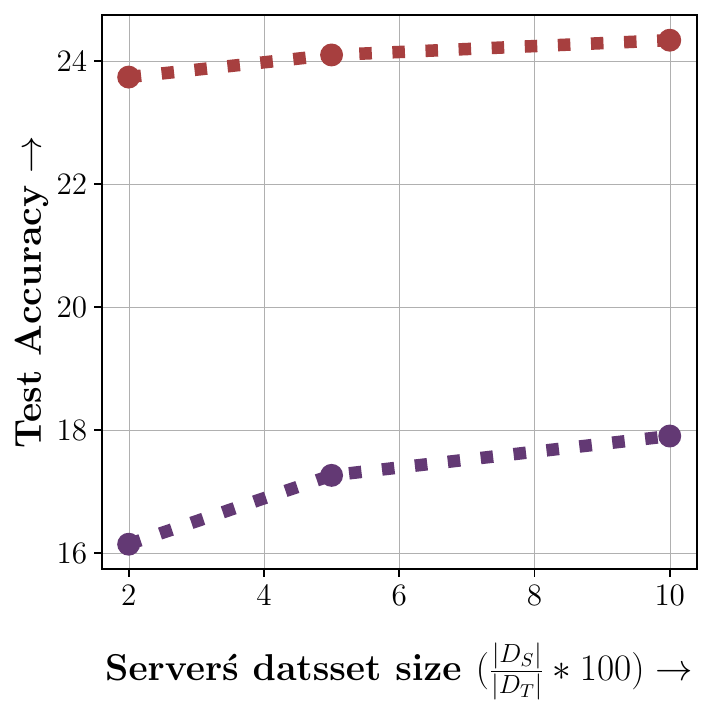}
\caption*{$\underbracket[1pt][1.0mm]{\hspace{3.0cm}}_{\substack{\vspace{-4.0mm}\\
\colorbox{white}{(a) \scriptsize CIFAR100}}}$}
\phantomcaption
\label{fig:CIFAR100}

\end{subfigure}
\begin{subfigure}[b]{0.39\linewidth}
\centering
\includegraphics[width=\linewidth]{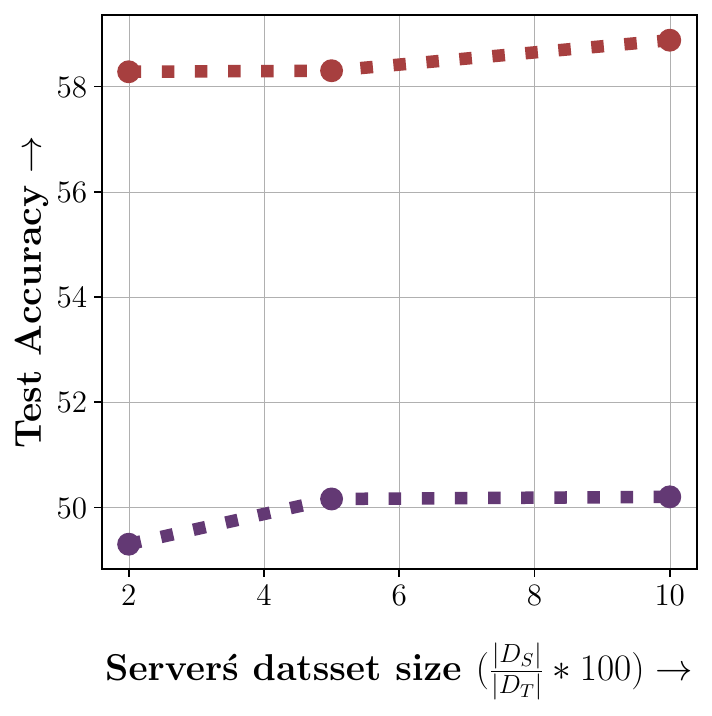}
\caption*{$\underbracket[1pt][1mm]{\hspace{3.0cm}}_{\substack{\vspace{-4.0mm}\\
\colorbox{white}{(b) \scriptsize CIFAR10}}}$}
\phantomcaption
\label{fig:CIFAR10}
\end{subfigure}
\caption{Impact of server's dataset size on \model{} performance under $20\%, 40\%$ close-set noise.}
\label{fig:val_variation}
\end{figure}

\subsection{Ablation on the size of $|\valdata|$} \label{sec:ab_ch_val}
Here, we investigate how the size of the server's validation dataset $D_S$ influences the coreset selection by clients. We examine the impact by setting $D_S$ to represent $2\%$, $5\%$, and $10\%$ of the samples from $D_T$. This analysis is conducted under conditions with both $20\%$ and $40\%$ closed-set label noise, using CIFAR10 and CIFAR100 datasets. The results, presented in Figure \ref{fig:val_variation}, demonstrate \model's consistent performance across varying $D_S$ sizes.

\section{Conclusion}

In this work, we introduced a new approach called \model\ to address the challenge of learning in a federated setting where the distribution of data across the client nodes is \noniid, and ingested with noise. Our proposed approach selects a coreset from each client that best approximates the server's last layer gradient. Our experimental results illustrate that \model\ outperforms state-of-the-art methods, and achieves the best accuracy \vs\ efficiency trade-off when the datasets are not noisy. In case of noise, \model\ was able to achieve significant gains compared to other FL and coreset selection baselines.

\section{Acknowledgements}
Durga Sivasubramanian and Ganesh Ramakrishnan express their gratitude to Google Research award for their assistance and funding. Ganesh Ramakrishnan is also grateful to the IIT Bombay Institute Chair Professorship for their support and sponsorship.


{\small
\bibliographystyle{ieee_fullname}
\bibliography{gcfl_main}
}
\clearpage
\newpage

\onecolumn
\leftline{ {\Large Appendix } }

\section{Code}
We have released the code in github: \url{https://github.com/nlokeshiisc/GCFL_Release/tree/master}.

\section{Notation}
We list the notations and their corresponding description used in \model{} in the Table \ref{tab:notations}.

\begin{table}[h]
    \centering
    \resizebox{0.7\linewidth}{!}{
    \begin{tabular}{c l}
    \hline \hline
       Notation & Description\\ \hline \hline \\
       $\cliset$ & Set of clients $\{c_1,c_2,\cdots, c_N\}$ \\
       $N$ & No. of clients \\
       $D_T$ & Training data partitioned across $N$   clients i.e. $\trndata = \bigcup_{i=1}^N D_i$ \\
       $D_i$ & Data at each client $i$\\
       $\valdata$ & Server's data \\
       $\coreset, w_i^t$ & Subset selected at client $i$ on round $t$  and its associated weights \\
        $\serverloss$ &  Server loss computed on $\valdata$ \\
       $\cliloss$ & Loss at each client \\
       $\eta_l$ & local learning rate at each client \\
       $\eta_g$ & global learning rate at the server \\
       $E$ & Denotes the number of local gradients update steps that the client  performs \\
        \hline
        \hline
    \end{tabular}}
    \caption{ Important Notations and Descriptions} 
    \label{tab:notations}
\end{table}

\section{GCFL Algorithm}

\begin{algorithm}[H]
\caption{\model{} algorithm~\label{alg}}
\label{alg:greedy}
\begin{algorithmic}[1]
\REQUIRE \small{ Clients $\mathcal{C} = \{c_1, \cdots, c_N\}$, Server $\mathcal{S}$, Training Data at client $i$ $D_i = \{(x_{ij}, y_{ij})_{j=1}^{n_i}\}_{i=1}^N$, Server Data $\valdata$, communication rounds $T$, budget $b$, number of clients per round $m$, local and global learning rates $\eta_l, \eta_g$, , local gradient steps $E$, rounds at which server gradient is broadcasted $K$}
    \STATE $\theta_0 \leftarrow \textsc{init\_}\flmdl$; \;\;$\Xcal_i^0 \leftarrow \textsc{init\_random\_samples}(i) \;\; \forall i \in [N]$
      \FOR{round $t \in [T]$}
            \STATE  \tikzmk{A} \textbf{server does:}
            \STATE $C^t \leftarrow$ sample $m$ out of $N$ clients
                \STATE Broadcast ($\theta^t, \nabla_\theta \serverloss(\valdata; \theta^t)$) if $t \% K = 0$ else Broadcast $\theta^t$ to $C^t$ \\
            \tikzmk{B} \boxit{pink} 
            \vspace{-0.1cm}
            \STATE \tikzmk{A} \textbf{each client $c_i \in C^t$ does:}
                \STATE $\coreset \leftarrow $  solve Eq. ~\eqref{eq:gmobj} using greedy algorithm if $t \% K = 0$ else  $\Xcal_i^{t-1}$
            \STATE Set $\theta^{'} \leftarrow \theta^t$
            \FOR{$e \in [E]$}
                \STATE sample a mini-batch $\Bcal \overset{\iid}{\sim} \coreset$
                \STATE $\theta^{'} \leftarrow \theta^{'} - \eta_l \frac{1}{|\Bcal|}\sum_{(x, y) \in \Bcal}\cliloss(f_{\theta^{'}}(x), y)$
            \ENDFOR 
            \STATE Broadcast $\delta_i^t \leftarrow \theta^t - \theta^{'}$ to $\mathcal{S}$ \\
            \tikzmk{B} \boxit{cyan}
            \vspace{-0.1cm}
            \STATE \tikzmk{A} \textbf{server does:}
            \STATE $\theta^{t+1} \leftarrow \theta^t + \eta_g \sum_{i \in S^t} \delta_i^t$  \\
            \tikzmk{B} \boxit{pink}
            \vspace{-0.1cm}
        \ENDFOR
    \RETURN Final model $f_{\theta^T}$
\end{algorithmic}
\end{algorithm}

\begin{algorithm}[H]
\caption{Coreset Selection Iteration for the $(k+1)^{\text{th}}$ Point}
\label{alg:grad_match}
\begin{algorithmic}[1]
    \REQUIRE Coreset $\ompsubset_i^k$, Weights $\corewt_i^k$, Budget $b$, Validation Gradient $\theta^t$
    \STATE Set $r^0 \leftarrow$  server's broadcasted validation gradient $\theta^t$ if $k=0$
    \STATE Calculate error residue: $r^k \leftarrow \sum_{j \in \ompsubset_i^k} \corewt_{ij}^k \nabla_\theta \cliloss^j(\theta^t) - r^{k-1}$\;
    
    \FOR{Data point $j \in [n_i] \setminus \ompsubset_i^k $}
        \STATE Calculate the distance: $d_j \leftarrow \big\Vert \nabla_{\theta}\ell_i^j(\theta^t) - r^{k}\big\Vert$\;
    \ENDFOR
    
    \STATE Select the data point minimizing distance: $j^\star \leftarrow \underset{j}{\argmin} {\;\; d_j}$\;    
    
    \RETURN \; Next data point $j^\star$
\end{algorithmic}
\end{algorithm}

Algorithm \ref{alg} presents the pseudocode of our proposed approach. The algorithm depicts the actions performed by both the server and clients at each round. The server's actions are are highlighted in orange and the clients' actions shown in blue color. 

At a communication round $t$, first the server samples $m$ clients (line 3), and then broadcasts $\theta^t$ to them. Additionally, if the clients need to perform coreset selection, the server also broadcasts the validation gradients. The clients subsequently use these to construct the coreset $\coreset$ iteratively by solving our label-wise OMP problem (line $8$ of the algorithm). Then clients use the selected coreset to train the model for $E$ epochs (line 10--13). Finally, the parameters of the updated model are uploaded back to the server (line $14$). The server waits until it receives the updated model from all the sampled clients $C^t$, then it averages them and the updates the global model with the average(line $17$). This completes one round of \model{} execution. 

\section{Datasets Details}
\begin{figure}[H]

\centering
\begin{subfigure}[b]{0.32\linewidth}
\centering
\includegraphics[width=\linewidth]{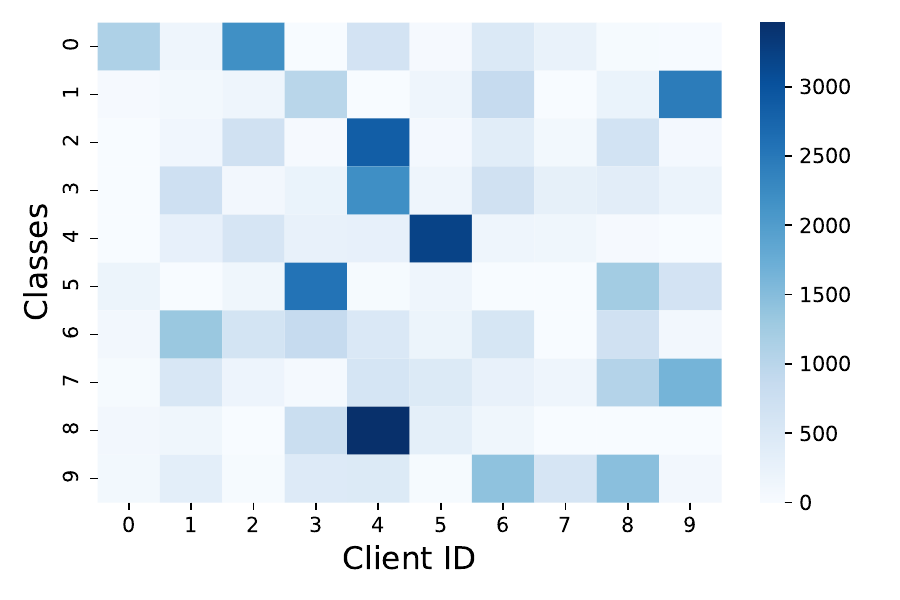}
\caption{CIFAR10} 
\end{subfigure}
\begin{subfigure}[b]{0.32\linewidth}
\centering
\includegraphics[width=\linewidth]{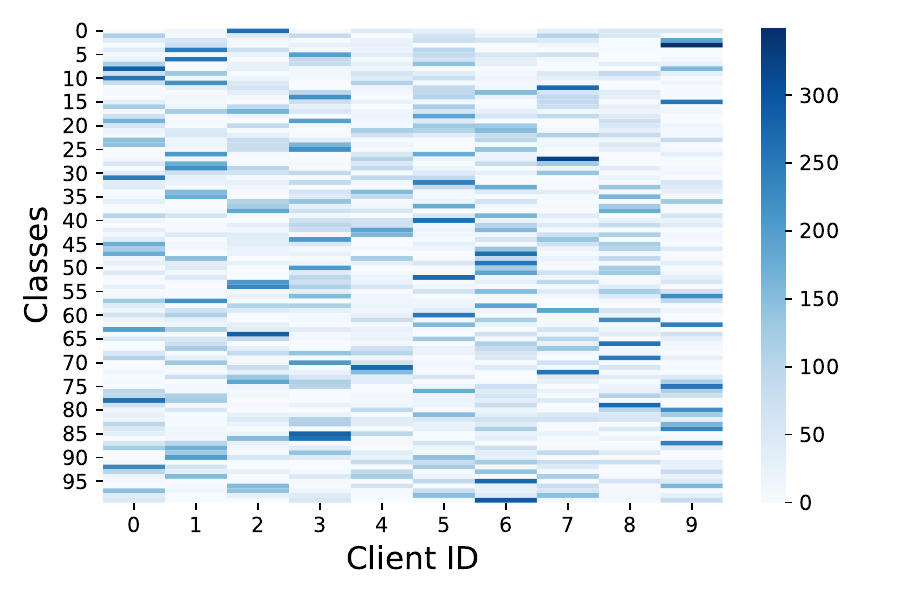}
\caption{CIFAR100} 
\end{subfigure}
\begin{subfigure}[b]{0.32\linewidth}
\centering
\includegraphics[width=\linewidth]{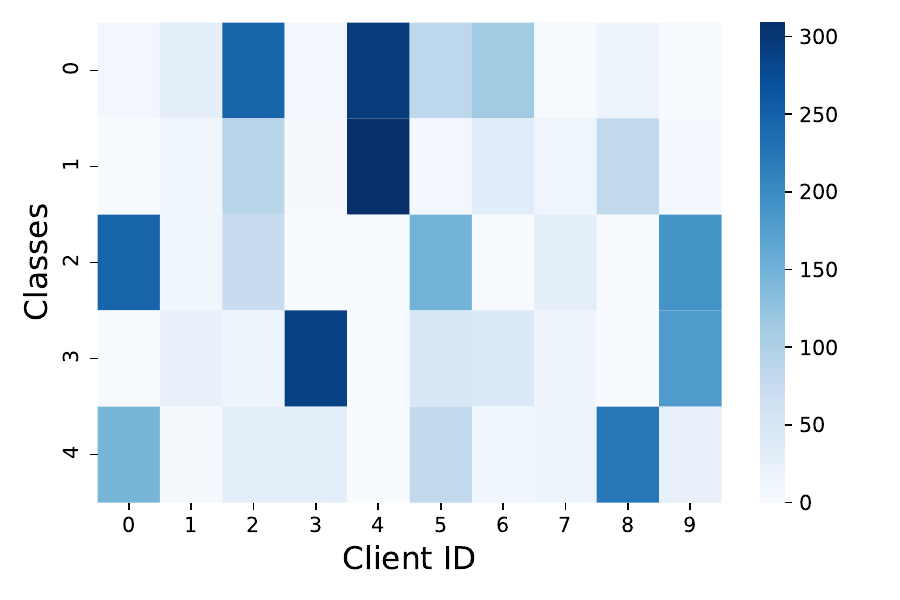}
\caption{FLOWERS} 
\end{subfigure}
\begin{subfigure}[b]{0.32\linewidth}
\centering
\includegraphics[width=\linewidth]{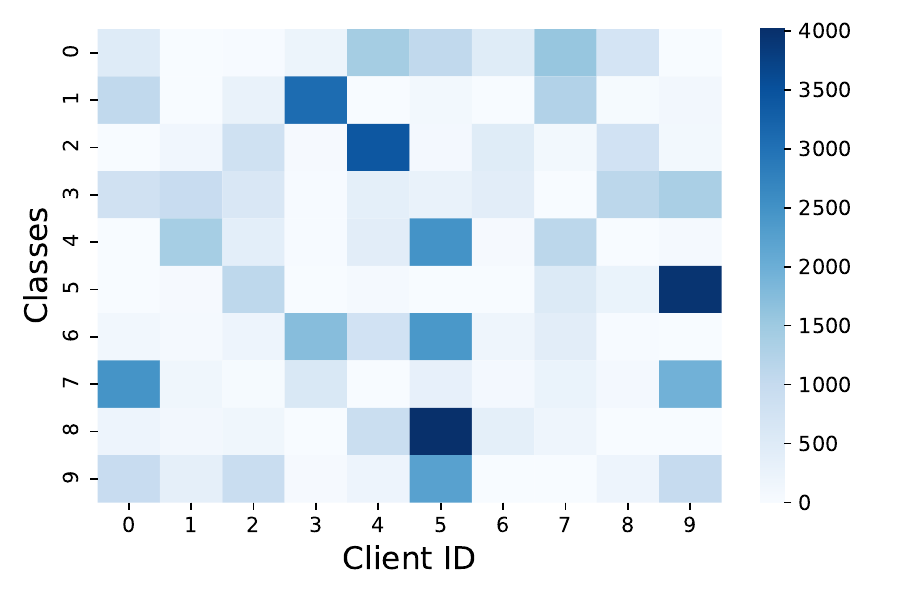}
\caption{FEMNIST} 
\end{subfigure}
\caption{Classwise distribution of training instances across clients. The X axis spans the clients and the Y axis spans the classes. Each rectangle represents the number of instances that belong to a particular a class in a client. A dark rectangle in the cell $(i, j)$ means that the $i^{\text{th}}$ client has more instances of class $j$. }
\label{fig:clinetDis} 
\end{figure}

In table \ref{tab:textdatasplits} we list the details about the datasets used in the experiments. Further, to showcase the challenging nature of datasets used in our experiments, we show the \noniid\ nature of the partition across clients using heatmaps in Figure \ref{fig:clinetDis}. We observe that each client is characterized by an allocation of data points $D_i$ such that it is skewed in favor of one or two classes. 

{\renewcommand{\arraystretch}{1.3}%
\begin{table}[h]
    \centering
    \resizebox{0.6\linewidth}{!}{
    \begin{tabular}{|l|c|c|c|c|c|}
    \hline
    \hline
        Dataset & \#Classes   & \#Clients' data & \#Server's data  & \#Test  \\ 
         &$(|\Ycal|)$  &  $(| \bigcup_{i =1}^N D_i|)$ & $(|\valdata|)$ &  $(|D_{\text{Test}}|)$ \\ \hline \hline
        Flowers & 5   & 3270  & 200  & 200    \\ \hline
        FEMNIST &  10  & 60000  & 5000   & 5000    \\ \hline
        CIFAR10 & 10 &   50000  & 5000   & 5000   \\ \hline
        CIFAR100 & 100 & 50000  & 5000   & 5000    \\ \hline
        \hline
    \end{tabular}}
    \caption{Dataset statistics} 
    \label{tab:textdatasplits}
\end{table}
}

\section{Additional experiments}


\subsection{Client Selection} \label{sec:ab_ch_num_p}

\begin{figure}[H]
\centering
\includegraphics[width =0.95\linewidth, trim={0 .5cm 0 0},clip] {IJCAI/new_figure/Legend.pdf}
\centering
\begin{subfigure}[b]{0.30\linewidth}
\centering
\includegraphics[width=\linewidth]{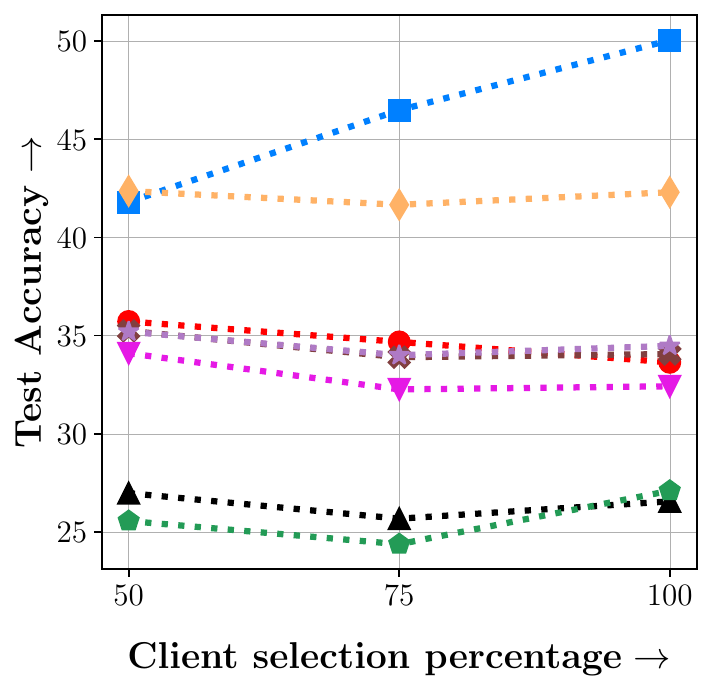}
\caption{10 clients}
\end{subfigure}
\begin{subfigure}[b]{0.30\linewidth}
\centering
\includegraphics[width=\linewidth]{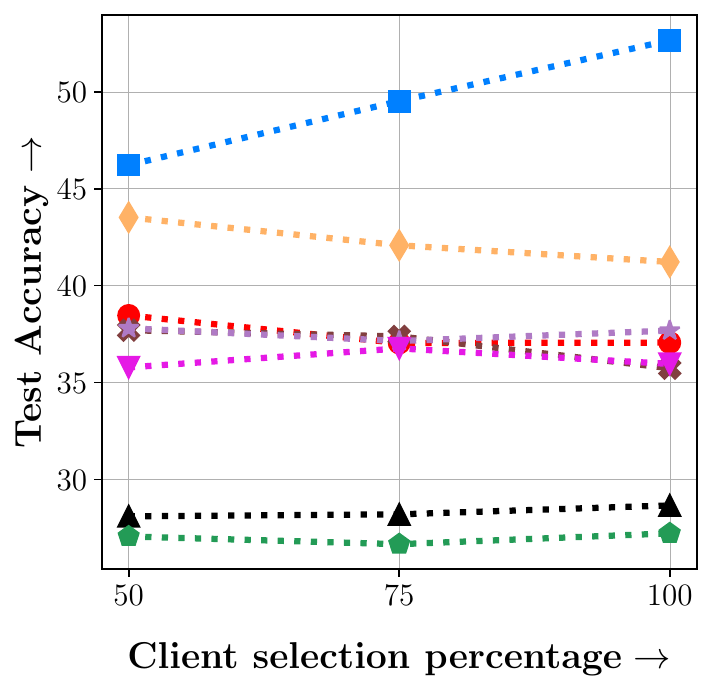}
\caption{15 clients}
\end{subfigure}
\begin{subfigure}[b]{0.30\linewidth}
\centering
\includegraphics[width=\linewidth]{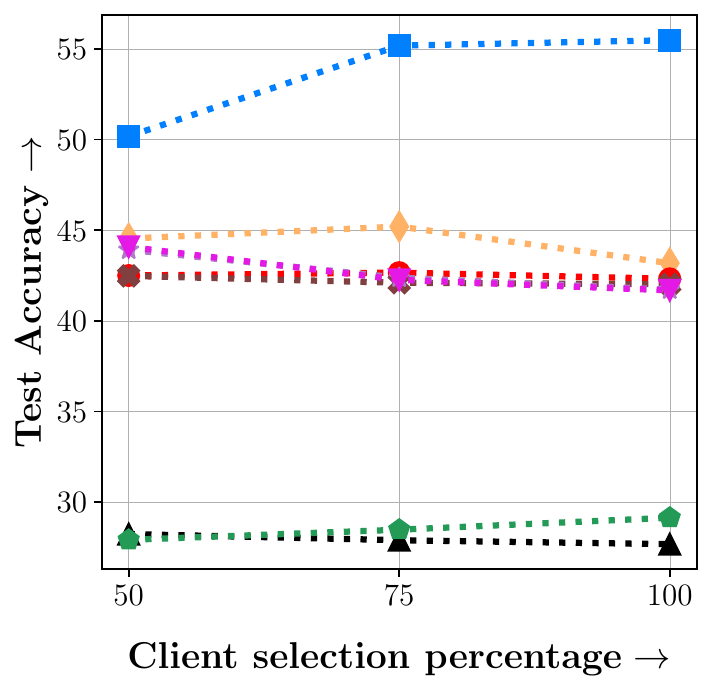}
\caption{25 clients}
\end{subfigure}

\caption{In this experiment we vary the number of participating clients $m$ in each round. We experiment with CIFAR-10 dataset that is injected with $40\%$ closed-set noise. Overall we observe that \model\ performs the best.}
\label{fig:client_closed}
\end{figure}

To test the robustness of \model{} in a setting where each communication round involves partial participation of clients, we conducted an experiment by varying the number of clients that are sampled in each round. In particular, we experimented with client participation ratios $50\%$, $75\%$ and $100\%$ and present the results on CIFAR10 dataset under $40\%$ closed-set noise setting. From the figure \ref{fig:client_closed} is it clear that the robustness of \model{} is not affected by the limited participation of clients.

\subsection{Ablation study: Varying Budget $b$} \label{sec:ab_var_bud}
\begin{figure}[h]
\centering
\includegraphics[width =0.15\linewidth] {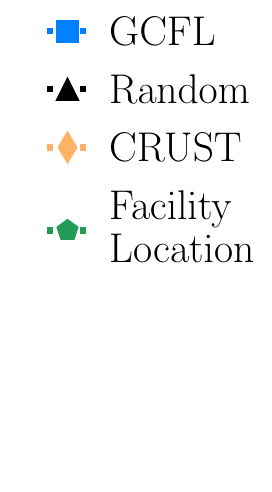}
\centering
\begin{subfigure}[b]{0.30\linewidth}
\centering
\includegraphics[width=\linewidth]{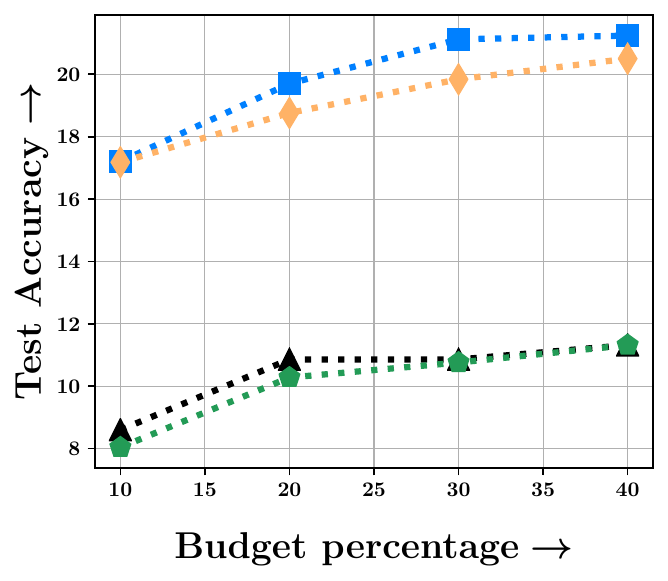}
\caption*{$\underbracket[1pt][1.0mm]{\hspace{3.25cm}}_{\substack{\vspace{-4.0mm}\\
\colorbox{white}{(a) \scriptsize CIFAR100}}}$}
\phantomcaption

\end{subfigure}
\begin{subfigure}[b]{0.3\linewidth}
\centering
\includegraphics[width=\linewidth]{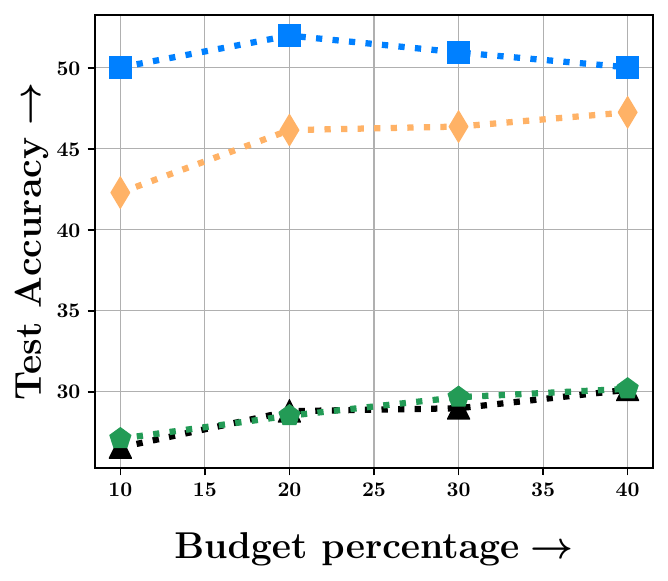}
\caption*{$\underbracket[1pt][1mm]{\hspace{3.25cm}}_{\substack{\vspace{-4.0mm}\\
\colorbox{white}{(b) \scriptsize CIFAR10}}}$}
\phantomcaption
\end{subfigure}
\caption{Performance of \model\ and other subset selection method as we vary budget in 40\% closeset noise setting.}
\label{fig:bugdet_experiments}
\end{figure}

All coreset selection algorithms' performances are affected by the size of the subset they are allowed to select. We present the effect of varying the sampling budget on coreset selection algorithms under $40\%$ close-set noise in Figure \ref{fig:bugdet_experiments}. For a fair comparison, we make the number of SGD steps consistent across different sampling budgets. For CIFAR10, in Figure \ref{fig:bugdet_experiments} a) we see that \model{}'s performance is robust to sampling budget and other coreset methods slightly improve as budget size increases. We attribute this robustness to the fact that \model{} selects coreset based on label-wise last layer server gradients. However, for CIFAR100 dataset, in Figure \ref{fig:bugdet_experiments} b) we see all the methods improve with increase in the budget size, since it is slightly a more difficult dataset having $100$ classes. 

\subsection{Ablation study: Varying \noniid-ness among clients $\alpha$} \label{sec:ab_var_niid}
\begin{figure}[h]
\centering
\includegraphics[width =0.15\linewidth,trim={.25cm 2cm 0 0}] {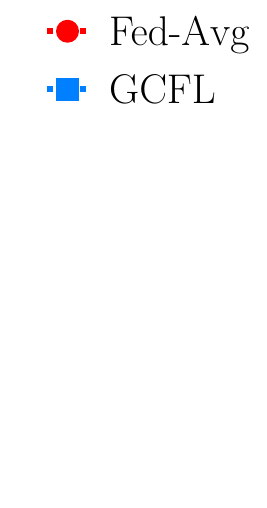}
\centering
\begin{subfigure}[b]{0.3\linewidth}
\centering
\includegraphics[width=\linewidth]{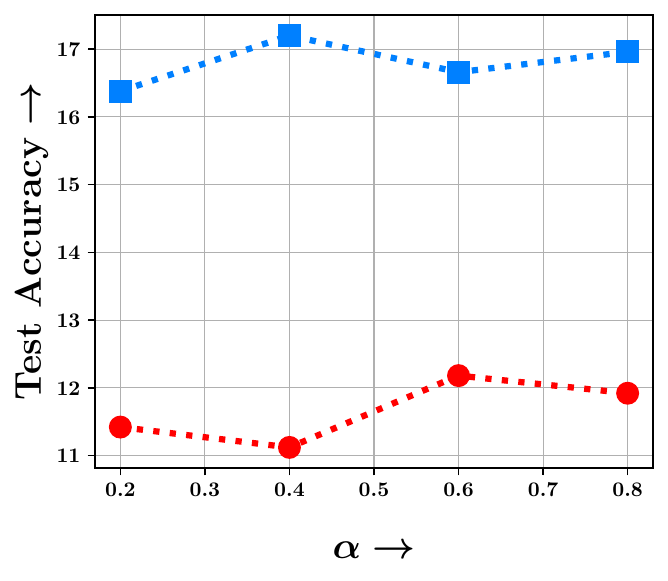}
\caption*{$\underbracket[1pt][1.0mm]{\hspace{3.25cm}}_{\substack{\vspace{-4.0mm}\\
\colorbox{white}{(a) \scriptsize CIFAR100}}}$}
\phantomcaption

\end{subfigure}
\begin{subfigure}[b]{0.3\linewidth}
\centering
\includegraphics[width=\linewidth]{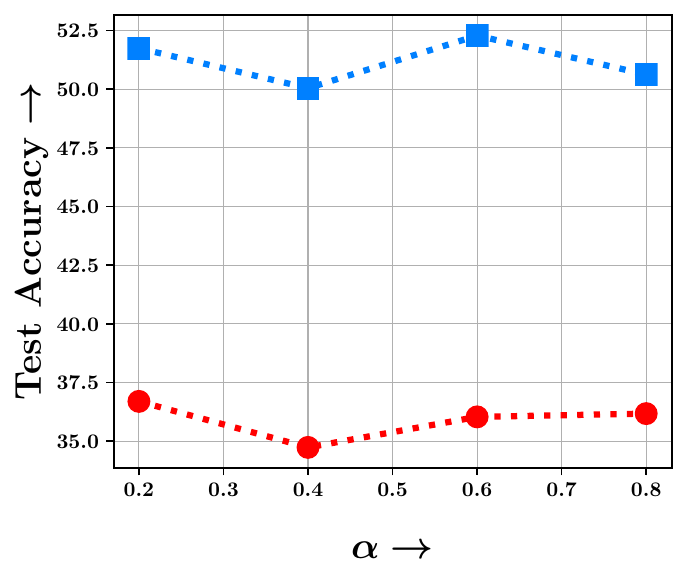}
\caption*{$\underbracket[1pt][1mm]{\hspace{3.25cm}}_{\substack{\vspace{-4.0mm}\\
\colorbox{white}{(b) \scriptsize CIFAR10}}}$}
\phantomcaption
\end{subfigure}
\caption{Performance of \model\ and Federated averaging as we vary $\alpha$(parameter controlling \noniid\  split among clients) in 40\% closeset noise setting.}
\label{fig:alpha_experiments}
\end{figure}

We distributed the data to clients following \cite{fedem} where we simulate the \noniid data partition by sampling class proportions from a symmetric Dirichlet Distribution parameterized with $\alpha$. Typically, setting a lesser $\alpha$ would result in a very skewed class proportion across clients, and as $\alpha$ increases, we reach the uniform partition (\ie\ \iid) in the limit. We conduct experiments under the closed set label noise with a noise ratio of $40\%$ to assess the performance of \model{} as against the standard Federated Averaging algorithm. The results are presented in Figure \ref{fig:alpha_experiments} which clearly elucidates the robustness of \model{} primarily attributed to its ability to balance the corset across classes by deliberately running the selection on a per-class basis alongside selecting noise-free informative points. On the other hand, Federated averaging, due to its sensitivity to noise, is unable to recover even when the partition reflects \iid\ characteristics.

\end{document}